\documentclass[letterpaper]{article} 
\usepackage{aaai2026}
\usepackage{times}  
\usepackage{helvet}  
\usepackage{courier}  
\usepackage[hyphens]{url}  
\usepackage{graphicx} 
\usepackage{natbib}  
\usepackage{caption} 
\usepackage{algorithm}
\usepackage{algorithmic}
\usepackage{amssymb}
\usepackage{amsmath}
\usepackage{booktabs}
\usepackage{makecell}
\usepackage{multirow}
\usepackage{xcolor}
\usepackage{pifont}
\usepackage{newfloat}
\usepackage{listings}
\DeclareCaptionStyle{ruled}{labelfont=normalfont,labelsep=colon,strut=off} 
\floatstyle{ruled}
\newfloat{listing}{tb}{lst}{}
\floatname{listing}{Listing}
\title{Federated Graph Unlearning}

\author{
    Yuming Ai\equalcontrib\textsuperscript{\rm 1},
    Xunkai Li\equalcontrib\textsuperscript{\rm 1},
    Jiaqi Chao\textsuperscript{\rm 2},
    Bowen Fan\textsuperscript{\rm 1},\\
    Zhengyu Wu\textsuperscript{\rm 1},
    Yinlin Zhu\textsuperscript{\rm 3},
    Rong-Hua Li\textsuperscript{\rm 1},
    Guoren Wang\textsuperscript{\rm 1}
}

\affiliations{
    \textsuperscript{\rm 1}Beijing Institute of Technology, Beijing, China\\
    \textsuperscript{\rm 2}Shandong University, Shandong, China\\
    \textsuperscript{\rm 3}Sun Yat-Sen University, Guangdong, China\\
    1120211787@bit.edu.cn, cs.xunkai.li@gmail.com, 202300800028@mail.sdu.edu.cn, \\
    3220241345@bit.edu.cn, jeremywzy96@outlook.com, zhuylin27@mail2.sysu.edu.cn, \\lironghuabit@126.com, wanggrbit@gmail.com
}

\begin{document}

\maketitle

\begin{abstract}

The demand for data privacy has led to the development of frameworks like Federated Graph Learning (FGL), which facilitate decentralized model training. However, a significant operational challenge in such systems is adhering to the right to be forgotten. This principle necessitates robust mechanisms for two distinct types of data removal: the selective erasure of specific entities and their associated knowledge from local subgraphs and the wholesale removal of a user's entire dataset and influence.
Existing methods often struggle to fully address both unlearning requirements, frequently resulting in incomplete data removal or the persistence of residual knowledge within the system. This work introduces a unified framework, conceived to provide a comprehensive solution to these challenges. 
The proposed framework employs a bifurcated strategy tailored to the specific unlearning request. For fine-grained Meta Unlearning, it uses prototype gradients to direct the initial local forgetting process, which is then refined by generating adversarial graphs to eliminate any remaining data traces among affected clients. In the case of complete client unlearning, the framework utilizes adversarial graph generation exclusively to purge the departed client's contributions from the remaining network.
Extensive experiments on multiple benchmark datasets validate the proposed approach. The framework achieves substantial improvements in model prediction accuracy across both client and meta-unlearning scenarios when compared to existing methods. Furthermore, additional studies confirm its utility as a plug-in module, where it materially enhances the predictive capabilities and unlearning effectiveness of other established methods.
\end{abstract}

\section{Introduction}

The growing adoption of graph neural networks (GNNs)~\cite{wu2020comprehensive} in domains like social networks~\cite{ijcai2018p142}, bioinformatics~\cite{qu2023app_gnn_bio2}, and recommendation systems~\cite{cai2023app_gnn_rec3} underscores the value of graph-structured data. However, real-world graphs are inherently distributed across clients and often contain sensitive information. Centralized graph learning methods, which require data aggregation, risk privacy breaches and violate regulations such as GDPR~\cite{GDPR} and CCPA~\cite{CCPA1}. Federated graph learning~\cite{he2021fedgraphnn} thus emerges, combining privacy-preserving computation with graph representation learning. This approach ensures regulatory compliance while mitigating performance degradation from data heterogeneity via cross-client graph modeling.

One urgent privacy-preserving compliance to meet is granting users' rights to be forgotten~\cite{yang2024machine}. However, most existing federated graph learning methods lack this capability, limiting their applicability in real-world deployments and introducing model bias~\cite{chang2023bias} and security vulnerabilities~\cite{tolpegin2020data}.
Unlike most existing Federated Unlearning methods~\cite{romandini2024federaser, zhong2025fused, zhao2024mode} that are primarily designed for independent and identically distributed (i.i.d.) non-graph data such as images or text, designing a unified framework for Federated Graph Unlearning (FGU) presents unique challenges.

Unlearning entities from graph-structured data requires not only removing the target node but also its connectivity with neighboring nodes, thereby disrupting the topological integrity. Moreover, due to the dependency of GNNs on information propagation, such structural disturbances can significantly degrade the quality of neighboring nodes' learned representations.
In the context of FGU, this work formally proposes two unlearn requests that align with the collaborative training architecture: 

\ding{117} \textbf{Meta Unlearning:} Request to remove specific entities and their associated structural information from a client’s local subgraph during the federated training process. This ensures that historical interactions of forgotten entities no longer contribute to cross-client knowledge transfer. 

\ding{117} \textbf{Client Unlearning:} Request to remove an entire client during the federated training process, ensuring that all associated entities and their interactions cease contributing to the global learning process. The knowledge originated from forgotten clients that infused to other models during the federated process should also be removed.

Formal mathematical representations of these two unlearn requests are provided in the next section. 
Both requests reflect common real-world demands for protecting user data privacy, such as requests to delete personal accounts or to fully withdraw a client from federated participation due to new regulatory restraints or compromised security.

\textbf{Challenge}. Although some prior work has attempted to address these challenges, key limitations persist: 

\ding{172} \textbf{Incomplete Unlearning Support}: Existing methods handle client unlearnin and meta unlearning in isolation, lacking a unified framework to coordinate multi-level unlearning. This limits their effectiveness in addressing heterogeneous unlearning requirements in practical settings. 

\ding{173} \textbf{Residual Knowledge Permeation}: During federated training, information from data marked for unlearning may propagate to other clients via gradient exchanges and graph-structural dependencies. Most prior approaches focus solely on the global model and the primary unlearning subject, neglecting cross-client contamination. Consequently, sensitive information might continue to be extracted.

\textbf{Method}.
To address these limitations, we propose PAGE, a novel three-stage unlearning framework. The first stage, Prototype Matching for Local Unlearn, efficiently removes a client's unique contributions by projecting its data prototype onto a shared feature subspace and minimally fine-tuning the model toward this common representation. Subsequently, the second stage, Adversarial Graph Generation, validates the process by creating adversarial inputs engineered to maximize the output discrepancy between the original and unlearned models, thereby acting as sensitive probes for any residual knowledge. Finally, if knowledge permeation is detected, the third stage, Negative Knowledge Distillation for Influenced Unlearn, identifies the influenced clients via prototypical similarity and applies a targeted distillation to eradicate the infiltrated information, ensuring a comprehensive and robust unlearning outcome.

\textbf{Contributions}.
(1) \textit{\underline{New Perspective}}.
This paper establishes a unified theoretical framework for federated graph unlearning that supports multiple unlearn requests. It sets the standard for achieving precise decoupling between forgotten and retained knowledge in federated graph scenarios for the first time.
(2) \textit{\underline{New Method}}.
We propose \textbf{PAGE}, a novel approach comprising three key components: a prototype matching module for guiding local unlearning, an adversarial graph generation module for verifying unlearning and mitigating residual impact, and a negative knowledge distillation module for removing unintended knowledge transfer.
(3) \textit{\underline{SOTA Performance}}.
Across 8 benchmark datasets, PAGE achieves state-of-the-art performance, improving prediction accuracy by 5.08\% in client unlearning and 1.50\% in meta-unlearning scenarios. More impressively, it yields up to 11.84\% improvement on large-scale graphs. PAGE also serves as an effective plug-in to enhance existing meta-unlearn methods as the ablation experiment reveals.


\section{Preliminaries}

\subsection{Federated Graph Learning}

In this subsection, we present a formal definition of federated graph learning (FGL). Assume that the FGL system comprises $K$ clients $\mathcal{C}=\{c^1,c^2,...,c^K\}$ and a central server. The global graph dataset is defined as $\mathcal{G}=\{\mathcal{G}^i|c^i \in \mathcal{C}\}$. The server retains no raw data but the global model parameter $\theta^{g}$. Each client $c^i$ holds a local subgraph $\mathcal{G}^i=\{\mathcal{V}^i,\mathcal{E}^i,\mathcal{X}^i\}$, where $\mathcal{V}^i$ represents the node set in $c^i$, $\mathcal{E}^i$ represents the edge set, $\mathcal{X}^i$ represents the feature set. In addition, each client holds a local model parameter $\theta^i$ with the same architecture as the global model. The objective of FGL is to collaboratively train the global graph neural network model $f_{\theta^{g}}:\mathcal{G}\rightarrow \mathcal{Y}$ (e.g., for node classification or link prediction) whose parameters $\theta^{g}$ are obtained by minimizing a global loss function:
\begin{equation}
    \min_{\theta^{g}} \mathcal{L}(\theta^{g}) = \sum_{i=1}^{K}\frac{|\mathcal{V}_i|}{|\mathcal{V}|} \cdot \mathcal{L}_i(\theta^{g};\mathcal{G}^i)
\end{equation}

For a more detailed exposition, we illustrate the training procedure of FGL at round $t$ using \textbf{FedAvg} as an example.

\textbf{Receive Message}.
Each client receives the global model parameters distributed by the server and initializes the local model through $\theta^i_t \leftarrow \theta^{g}_t$.

\textbf{Local Update}.
Each client performs multiple rounds of local training using its own subgraph. The goal of local training is to obtain $\theta^i_{t+1}$ by optimizing the local loss function:
\begin{equation}
    \theta^i_{t+1} = \min_{\theta^i_t} \mathcal{L}_{task}(\theta^i_t;\mathcal{G}^i)
\end{equation}

\textbf{Upload Message}.
Each client uploads the local model $\theta^i_{t+1}$ and sampled data size $n_i$ to the server.

\textbf{Global Aggregation}.
The server collects the parameters uploaded by all $K$ clients and performs weighted averaging:
\begin{equation}
    \theta^{g}_{t+1} = \sum_{i=1}^K \frac{n_i}{\sum_j n_j} \cdot \theta^i_{t+1}
\end{equation}

\subsection{Federated Graph Unlearning}
\label{subsec:Federated Graph Unlearning}
In this subsection, we give a formal definition of various unlearn requests and FGU.
The unlearn requests are divided into two categories: meta unlearning and client unlearning.

\textbf{Meta Unlearning}.
Meta unlearning represents the unlearning of the internal structure of the client subgraph. For example, in a social network scenario, if a user asks the platform to delete his account, the user node and all its associated edges need to be removed from the global social graph, and subsequent recommendation models no longer rely on their historical interactions. When client $c^i$ receiving a meta unlearn request $\Delta \mathcal{G}^i = \{\Delta \mathcal{V}^i,\Delta \mathcal{E}^i,\Delta \mathcal{X}^i \}$, it can be decomposed into three dimensions of unlearning: node level $\Delta \mathcal{G}^i = \{\Delta \mathcal{V}^i, \varnothing, \varnothing\}$, edge level $\Delta \mathcal{G}^i = \{\varnothing, \Delta \mathcal{E}^i, \varnothing\}$ and feature level $\Delta \mathcal{G}^i = \{ \varnothing, \varnothing, \Delta \mathcal{X}^i\}$. For each client, three dimensions of meta unlearn request may occur. Therefore, for the entire federated system, the total meta unlearn request is defined as $\Delta \mathcal{G} = \{\Delta \mathcal{V},\Delta \mathcal{E},\Delta \mathcal{X} \} = \bigcup_{i=1}^{K} \Delta \mathcal{G}^i$.

\textbf{Client Unlearning}.
Client unlearning means that the entire client exits the FGL system, requiring the removal of all its data contributions. For instance, if a bank withdraws from collaboration due to policy restrictions, all related account nodes and associated transaction edges must be removed from the global transaction graph without affecting the performance of the risk control model of other banks. When receiving a client unlearn request $\Delta \mathcal{C} = \{ c^{u_1},c^{u_2},...,c^{u_M}\}$, the FGL system needs to completely and verifiably remove all data contributions from some specific clients.

Existing graph unlearning (GU) methods support meta unlearn requests: GIF~\cite{wu2023gif} and IDEA~\cite{2024IDEA} quantify first-order deletion impacts via closed-form parameter adjustments, while D2DGN~\cite{2024D2DGN} and MEGU~\cite{xkliMEGU2024} balance unlearning and reasoning through specialized loss functions. For knowledge graphs, FedLU~\cite{zhu2023fedlu} eliminates triplet knowledge with neuroscience-inspired global propagation, and FedDM~\cite{liu2024feddm} leverages diffusion models to mitigate knowledge impacts via noisy data generation. Correspondingly, most federated unlearning methods only address client unlearn requests: FedEraser~\cite{romandini2024federaser} reconstructs models via historical update calibration; FUSED~\cite{zhong2025fused} trains sparse adapters for knowledge overwriting; MoDe~\cite{zhao2024mode} employs staged momentum degradation and distillation; ReGEnUnlearn~\cite{liu2025subgraph} reduces subgraph interference through optimal sampling and client-specific knowledge extraction.

\textbf{Notions}.
Assume that $\hat \theta$ is the randomly initialized original parameters As defined above, the FGL system obtains model parameters based on the client set $\mathcal{C}$ and the graph dataset $\mathcal{G}$ by $\theta^o = FGL(\hat \theta;\mathcal{C};\mathcal{G})$. When an unlearn request $ <\Delta \mathcal{G},\Delta \mathcal{C}>$ is received, the retrained model parameters $\theta^*$ are trained from scratch on retained graph $\mathcal{G}_r=\mathcal{G} \setminus \Delta \mathcal{G}$ and retained client set $\mathcal{C}_r=\mathcal{C} \setminus \Delta \mathcal{C}$ by $\theta^*=FGL(\hat \theta; \mathcal{C}_r; \mathcal{G}_r)$. The parameters of the unlearned model, denoted $\bar \theta$ are obtained by applying the FGU algorithm to the original model parameters $\theta^o$, defined as: $\bar \theta=FGU(\theta^o;\mathcal{C};\mathcal{G};\Delta \mathcal{C};\Delta\mathcal{G})$.

\textbf{Objective}.
From a mathematical definition, the goal of the FGU algorithm is to minimize the difference between $\bar \theta$ and $\theta^*$. From a practical standpoint, the fundamental goal of a FGU algorithm is to, under the strict requirements of privacy compliance, efficiently and completely eliminate the influence of specified data elements (nodes, edges, subgraphs, or entire clients) on the already‑trained global graph model, while simultaneously minimizing any degradation in the model’s ability to represent the remaining data—thereby striking an optimal balance between precise unlearning and model preservation.

\textbf{Unlearning Verification}.
Membership inference attack is a privacy-stealing technique for machine learning models, the core goal of which is to determine whether a specific data record is used in the training set of the target model. The attacker builds an inference model to distinguish between member data (training data) and non-member data (non-training data) by analyzing the model's response to input data (such as prediction confidence, gradient update, or recommendation sequence).

\begin{figure*}[t]

  \includegraphics[width=\textwidth]{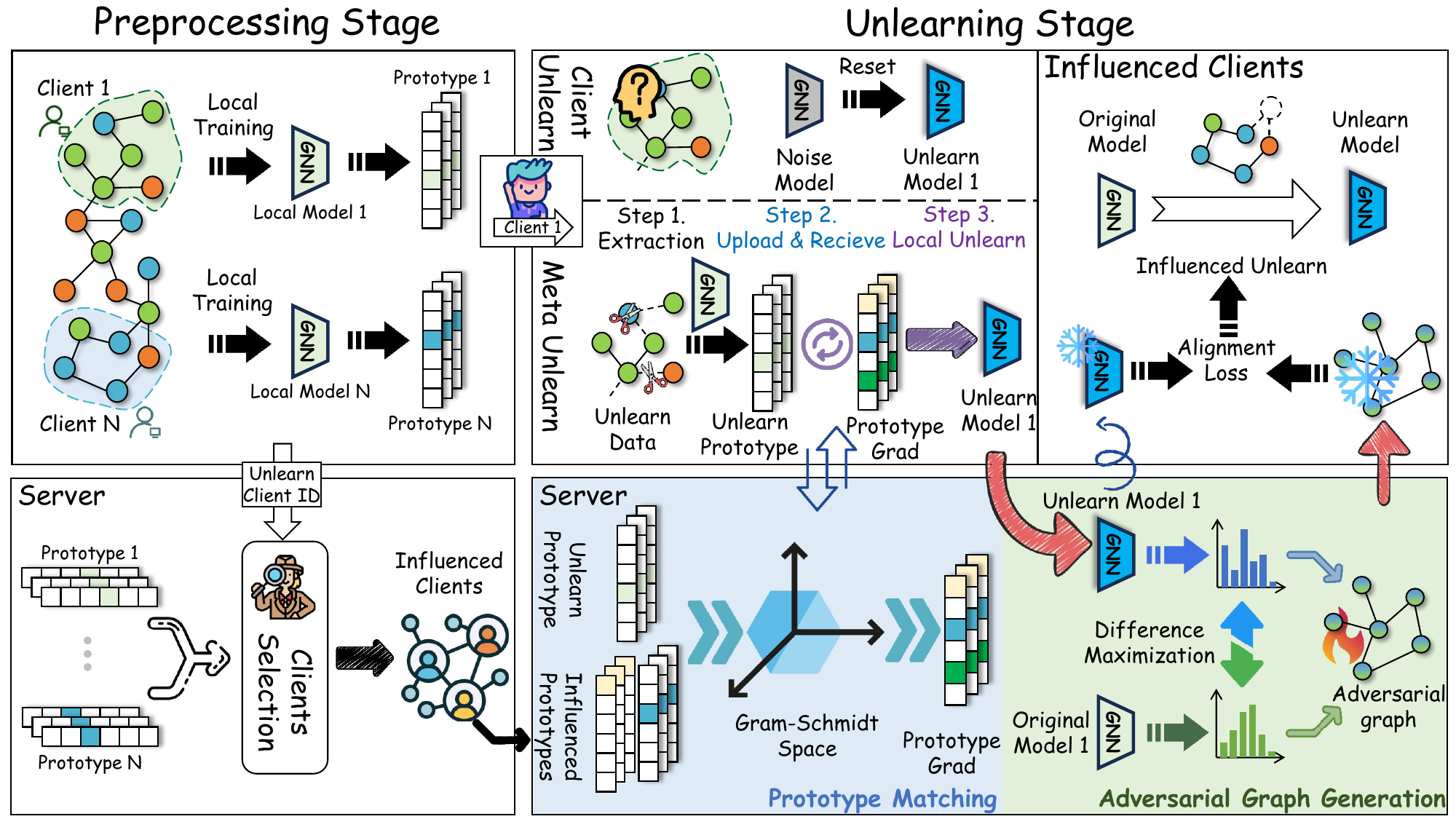}
  \caption{
  Overview of our proposed PAGE. Left: preprocessing before unlearning. Right: unlearning stage. The server performs prototype matching to guide the client to local unlearn. Based on the original model and unlearned model, adversarial graph samples are generated to guide the influenced client to perform influenced unlearn.
  }
  \label{fig: framework}
\end{figure*}

\section{Proposed Method}

In this section, we introduce PAGE, designed to address federated graph unlearning under multiple unlearn requests.
The methodology comprises three core steps:
(1) Prototype Matching for Local Unlearn: The server leverages prototypes to define unlearning objectives, guiding clients in the precise erasure of local knowledge;
(2) Adversarial Graph Generation: Constructing negative samples implicitly containing knowledge for unlearning by maximizing model discrepancies before and after unlearning;
(3) Negative Knowledge Distillation for Influence Unlearn: Eliminating the impact of knowledge permeation on associated clients.


\subsection{Architecture Overview}

As shown in Figure \ref{fig: framework}, in the preprocessing stage on the left, each client calculates a local prototype vector and uploads it to the server. The server calculates the cosine similarity between the vectors and filters out the influenced clients based on the similarity. On the right, PAGE operates in three stages. First, it performs precise local unlearning by leveraging semantic prototypes, enabling the server to isolate private data components based on client-provided feature centroids. Next, it validates the unlearning efficacy by generating adversarial graph samples that maximize the discrepancy between the model's pre- and post-unlearning states, thereby creating a sensitive probe for residual knowledge. Finally, the method addresses federated knowledge leakage by using prototype similarity to identify influenced clients and then applying negative knowledge distillation—guided by the adversarial samples—to approximate the desired unlearned state.

It is worth noting that under the client unlearn request, there is no need to perform prototype matching for local unlearning. When all of the data in the client need to be forgotten, the noise model $\hat \theta$, which is initialized randomly, represents  the unlearned model $\bar \theta^{u}$. Adversarial graph generation module and negative knowledge distillation module together form the influenced unlearning component.

\subsection{Prototype Matching}

The topological correlations in graph data cause conventional parameter-level unlearning to disrupt structural semantics, while federated learning's global-local knowledge coupling compounds these difficulties. We thus propose a prototype matching-based local unlearning mechanism with core principles:
(1) Representing data distributions abstractly via class prototype vectors (feature-space centroids), avoiding direct raw graph manipulation;
(2) Generalizing gradient matching's behavioral constraint concept to feature-space prototype alignment, reducing computation costs while preserving graph structure modeling.

\textbf{Prototype}.
The client $c^i$ trains a graph neural network $f_{\theta^i}$ on the local dataset $\mathcal{G}^i$ and calculates its category prototype vector $p^i$. Taking the node classification task as an example, for category $c$, the prototype is defined as the mean representation of the nodes of this category in the feature space:
\begin{equation}
    p^i_c = \frac{1}{|\mathcal{V}^i|} \sum_{(x^i,y^i=c)} f_{\theta^i}(x^i)
\end{equation}

where $\mathcal{V}^i$ is the set of all nodes of the category $c$ in the client $c^i$.
Then, all clients upload their model parameters $\{ \theta^i|i=1,2,...,K\}$ and prototype sets $\mathcal{P}$ to the server.
\begin{equation}
    \mathcal{P}=\{ p^i_c | i=1,2,...,K;c \in C\}
\end{equation}

The client $c^{u}$ computes the unlearn prototype $p^{del}$ derived from the unlearn data $\Delta \mathcal{G}^{u}$:
\begin{equation}
    p^{del}= \{\frac{1}{|\Delta \mathcal{V}^{u}|} \sum_{(x^{u},y^{u}=c)} f_{\theta^{u}}(x^{u}) |c \in C\}
\end{equation}

\textbf{Gram-Schmidt Prototype Space}.
The server aggregates prototype sets $\mathcal{P}$ from all non-unlearning clients. To eliminate redundant information and construct an orthogonal basis space, Gram-Schmidt orthogonalization is employed on the prototype vectors $\mathcal{P}$:
$v_1 = p_1/||p_1||_2, \space v_i = p_i - \sum_{j=1}^{i-1}<p_i,v_j>v_j, \tilde v_i = v_i/ ||v_i||_2 $

Subsequently, we obtain the orthogonal basis matrix $V=[\tilde v_1, \tilde v_2,..., \tilde v_M] \in \mathbb{R}^{d \times M},M<<N$, which spans the global prototype space $Span(\mathcal{P})$. Then, orthogonal projection decomposition is performed to project the unlearning prototypes $p^{del}$ onto $Span(\mathcal{P})$.
\begin{equation}
    p_{com} = VV^Tp^{del} = \sum_{i=1}^M <p^{del},\tilde v_i>\tilde v_i
\end{equation}

We can obtain the prototype grad $p_{priv} = p_{com} - p^{del}$, which satisfies the orthogonality.
\begin{equation}
    <p_{priv},\tilde v_i>=0, \forall i \in \{1,2,...,M \}
\end{equation}

This prototype grad $p_{priv}$ is termed the pure private knowledge guiding vector. Then, distribute $p_{priv}$ to the client $c^f$ to perform local unlearn.

\textbf{Local Unlearn}.
The client $c^f$ performs local unlearn based on the prototype grad $p_{priv}$:
\begin{equation}
    \mathcal{L} = MSE(p^{del},p_{priv})
\end{equation}

And then, we can get the unlearned model paramaters $\bar \theta^{u}$.

\subsection{Adversarial Graph Generation}

In federated graph unlearning, relying solely on local unlearn operations cannot adequately verify knowledge elimination effectiveness. The primary reasons are:
1. Concealment of Residual Knowledge: Unlearning may only be effective at specific decision boundaries, requiring sensitive samples to expose model behavioral changes;
2. Need for Cross-Client Impact Quantification: Knowledge permeation levels demand quantifiable assessment tools to provide precise unlearning guidance for other influenced clients.

To address this, we propose a difference-maximized graph adversarial generation method. Its core principle systematically explores regions where model behavior changes most significantly after unlearning by simultaneously perturbing graph structures and node features.

First, the server loads the original model parameters and unlearned model parameters $\theta^{u}, \bar \theta^{u}$ of the target unlearning client $c^u$ to establish a dual-model comparison framework. The initial inputs for adversarial graph generation comprise the node feature matrix $X_{init} \in \mathbb{R}^{N \times d}$ and adjacency matrix $A_{init} \in \{0,1\}^{N \times N}$, which are initialized randomly.

Subsequently, gradient ascent is employed to optimize perturbations in continuous space, with differentiable adjacency matrix weights $A_{var}$ and feature matrix $X_{var}$ defined as optimization variables. Refer to Algorithm \ref{alg: adv graph gen} for the specific optimization process.

\begin{algorithm}[t]
\caption{Adversarial Graph Generation.}
\label{alg: adv graph gen}
\begin{algorithmic}[1]
\STATE \textbf{Input:} Initial graph $\mathcal{G}_{init} = (A_{init}, X_{init})$, models $f_{\theta^{u}}$, $f_{\bar \theta^{u}}$, hyperparameters $\lambda=0.1$, $\epsilon_x=0.1$
\STATE \textbf{Output:} Adversarial graph $\mathcal{G}_{adv}$
\STATE Initialize optimization variables $A_{var}$, $X_{var}$
\WHILE{not converged}
    \STATE $\tilde{A} \gets \sigma(A_{var})$, $A_{sym} \gets (\tilde{A} + \tilde{A}^T)/2$
    
    \STATE $\mathcal{G} \gets (X_{var}, A_{sym})$
    \STATE $\mathcal{L}_{diff} \gets \text{CE}(f_{\theta^{u}}(\mathcal{G}), f_{\bar \theta^{u}}(\mathcal{G}))$
    
    \STATE $\mathcal{L}_{reg} \gets \| A_{sym} - A_{init} \|_1$
    
    \STATE $\mathcal{L}_{adv} \gets \mathcal{L}_{diff} - \lambda \mathcal{L}_{reg}$
    \STATE Update $A_{var}$, $X_{var}$ 
    
    \STATE $X_{var} \gets X_{init} + \text{clip}(X_{var} - X_{init}, -\epsilon_x, \epsilon_x)$
    
\ENDWHILE

\STATE \textbf{Post-processing:}
\STATE $\Delta A \gets \text{TopK}(|A_{var} - A_{init}|), K=5$
\STATE Construct $A_{final}$: 
\[
A_{final}[i,j] = 
\begin{cases} 
1 - A_{init}[i,j] & \forall (i,j) \in \Delta A \\
A_{var}[i,j] & \text{otherwise}
\end{cases}
\]
\STATE $X_{final} \gets X_{init} + \text{clip}(X_{var} - X_{init}, -\epsilon_x, \epsilon_x)$
\STATE $\mathcal{G}_{adv} \gets \left( X_{final},\  \{(i,j) \mid A_{final}[i,j] > 0.5\} \right)$
\STATE \textbf{Return} $\mathcal{G}_{adv}$
\end{algorithmic}
\end{algorithm}

\subsection{Negative Knowledge Distillation}

To address the challenge of knowledge permeation, this module proposes a negative knowledge distillation mechanism. Using graph adversarial samples, it compels affected clients to conform to the post-unlearning state in sensitive regions, thereby achieving targeted elimination of knowledge permeation.

The server broadcasts the unlearned model parameters $\bar \theta^{u}$ and graph adversarial samples $\mathcal{G}_{adv}$ to other affected clients. The output of the unlearned model on these graph adversarial samples represents the negative knowledge requiring elimination. In addition, in each affected client, the local positive knowledge needs to be preserved is obtained from the local model parameters $\theta^{i}$ and local data $\mathcal{G}^i$. Through this module we can get unlearned model parameters of each client $\bar \theta^{i}$.

\textbf{Distillation architecture}.
To preserve positive knowledge, we want to minimize the loss between $\bar \theta^{i}$ and $\theta^{i}$ for the local dataset $\mathcal{G}^i$:
\begin{equation}
    \mathcal{L}_{pos} = \mathcal{L}_{task}(f_{\bar \theta^{i}}(\mathcal{G}^i),f_{\theta^{i}}(\mathcal{G}^i))
\end{equation}

Similarly, in order to unlearn the knowledge requiring elimination, we hope that the loss between $\bar \theta^{i}$ and $\bar \theta^{u}$ on $\mathcal{G}_{adv}$ is as small as possible:
\begin{equation}
    \mathcal{L}_{neg} = \mathcal{L}(f_{\bar \theta^{i}}(\mathcal{G}_{adv}),f_{\bar \theta^{u}}(\mathcal{G}_{adv}))
\end{equation}

where, $\mathcal{L}$ is the chosen distillation measure, which is the Mean Squared Error (MSE) loss function in this paper.

Finally, we can get the final objective function:
\begin{equation}
    \mathcal{L} = \mathcal{L}_{pos} + \alpha \mathcal{L}_{neg}
\end{equation}

where $\alpha$ is the regularization hyperparameter that balances the trade-off between the effects of the two separators.

\section{Experiments}

\begin{table*}[]
\caption{Performance Comparison. ACC ± STD(\%) for node classification. The highest results are highlighted in \textbf{bold}.}
  \centering
  \resizebox{\textwidth}{!}{
    \begin{tabular}{c|cc|cc|cc|cc}
    \toprule[0.16em]
          & \multicolumn{2}{c|}{\textit{5 Clients}} & \multicolumn{2}{c|}{\textit{10 Clients}} & 
          \multicolumn{2}{c|}{\textit{15 Clients}} &
          \multicolumn{2}{c}{\textit{20 Clients}} \\
\cmidrule{2-9}          
& Cora & PubMed & CS & Photo &Tolokers &Minesweeper & Amazon-ratings & ogbn-arxiv \\
\cmidrule{2-9}          & \multicolumn{8}{c}{\textit{Client Unlearning (Unlearn Ratio=0.2)}} \\
    \midrule
    Retrain    &83.98 ± 1.52 &83.90 ± 0.82 &82.18 ± 0.43 &78.41 ± 1.37 &78.52 ± 1.51 &79.88 ± 0.11 &41.09 ± 0.48 &57.04 ± 0.28  \\
    
    FedEraser    &83.85 ± 2.37 &78.52 ± 3.21 &78.22 ± 1.46 &54.66 ± 4.72 &78.22 ± 1.61 &79.81 ± 0.26 &36.89 ± 0.40 &28.63 ± 2.84   \\
    FUSED   &49.43 ± 2.78 &83.17 ± 0.40 &72.68 ± 4.17&67.60 ± 5.34 &78.47 ± 1.54 &79.86 ± 0.10 &39.08 ± 0.48 &38.29 ± 0.43   \\
   MoDe   &80.33 ± 4.04 &83.20 ± 1.13 &73.90 ± 3.42 &42.47 ± 8.23 &78.47 ± 1.54 &79.86 ± 0.10 &36.63 ± 4.19 &28.31 ± 3.91   \\

   ReGEnUnlearn  &83.83 ± 1.29 &\textbf{83.74 ± 0.95} &78.27 ± 1.92 &58.43 ± 9.78 &\textbf{78.54 ± 1.51} &80.04 ± 0.21 &40.20 ± 1.04 & 21.59 ± 5.30  \\
    Ours  &\textbf{83.91 ± 1.29} &83.57 ± 0.54 &\textbf{78.66 ± 1.06} &\textbf{69.73 ± 2.16} &78.52 ± 1.49 &\textbf{80.06 ± 0.04} &\textbf{40.75 ± 0.58} & \textbf{50.13 ± 2.24} \\
    \midrule
          & \multicolumn{8}{c}{\textit{Meta Unlearning (Unlearn Ratio=0.1)}} \\
    \midrule
    Retrain    &78.65 ± 0.04 &83.91 ± 0.17 &87.74 ± 0.09 &88.62 ± 0.13 &78.52 ± 0.18 &79.90 ± 0.10 &46.58 ± 0.06 &69.37 ± 0.02   \\
    GIF    &81.02 ± 0.38 &84.21 ± 0.13 &87.67 ± 0.05 &80.44 ± 0.95 &78.12 ± 0.08 &79.69 ± 0.33 &45.62 ± 0.13 &67.37 ± 0.02   \\
    D2DGN   &75.83 ± 0.61 &83.23 ± 0.15 &86.98 ± 0.33 &79.69 ± 3.53 &77.49 ± 0.26 &76.14 ± 1.16 &45.74 ± 0.20 &55.19 ± 0.05    \\
   MEGU  &77.96 ± 3.80 &83.43 ± 0.05 &86.94 ± 0.32 &82.03 ± 3.03 &78.34 ± 0.04 &79.87 ± 0.00 &42.63 ± 0.28 & 56.11 ± 0.13  \\
    FedLU  &71.96 ± 0.43 &73.14 ± 0.17 &80.44 ± 0.16 &80.21 ± 0.51 &78.32 ± 0.04 &79.87 ± 0.00 &44.11 ± 0.09 &62.75 ± 0.02  \\
    FedDM  &80.11 ± 0.63 &80.89 ± 0.20 &80.67 ± 3.54 &51.87 ± 2.57 &76.82 ± 1.37 &79.89 ± 0.08 &32.09 ± 1.23 &14.68 ± 3.03  \\
    Ours  &\textbf{81.29 ± 0.17} & \textbf{84.39 ± 0.09} &\textbf{88.66 ± 0.07} &\textbf{88.84 ± 0.39} &\textbf{78.69 ± 0.19} &\textbf{79.92 ± 0.10} &\textbf{46.21 ± 0.13} &\textbf{68.19 ± 0.08}  \\
    \bottomrule[0.16em]
    \end{tabular}}

  \label{Q1}%
\end{table*}%

\begin{table}[t]
\caption{Performance with edge and feature unlearning.}
\resizebox{\linewidth}{24mm}{
\setlength{\tabcolsep}{2mm}{
\begin{tabular}{cc|cccc}
\midrule[0.3pt]
\multirow{2}{*}{\makecell{Simulation \\Mode}}    & \multirow{2}{*}{Strategy} & \multicolumn{2}{c}{Physics} & \multicolumn{2}{c}{Computers} \\
                             &                           & Edge-Level    & Feature-Level  & Edge-Level      & Feature-Level   \\ \midrule[0.3pt]
\multirow{7}{*}{Metis}
& Retrain  
&92.99 ± 0.07 &93.12 ± 0.16 &87.09 ± 0.22  &84.07 ± 4.47     \\
& GIF 
&92.63 ± 0.03 &93.01 ± 0.07 &72.10 ± 4.79  &71.96 ± 3.86       \\
& D2DGN 
&88.12 ± 0.17 &89.01 ± 0.28 &71.51 ± 0.67  &72.34 ± 1.63      \\
& MEGU                      
& 92.59 ± 0.05 & 92.60 ± 0.03 & 73.01 ± 1.18 & 68.17 ± 4.91   \\
& FedLU
&91.15 ± 0.03 &89.74 ± 0.78 &79.89 ± 1.27 &79.66 ± 0.19 \\
& FedDM
& 77.28 ± 1.35&88.94 ± 1.90 &45.88 ± 8.11 &53.83 ± 0.59 \\
& PAGE
&\textbf{93.40 ± 0.03} &\textbf{93.52 ± 0.03} &\textbf{82.04 ± 0.43} &\textbf{82.42 ± 0.06} \\
\midrule[0.3pt]
\multirow{7}{*}{\makecell{Metis-Plus}}  
& Retrain  
&92.39 ± 0.01 &92.70 ± 0.07 &86.95 ± 0.23  &86.07 ± 0.69     \\
& GIF 
&90.37 ± 2.73 &92.45 ± 0.18 &74.18 ± 4.78  &66.84 ± 6.14       \\
& D2DGN 
&91.17 ± 0.18 &84.95 ± 6.05 &80.21 ± 0.53  &80.26 ± 0.02      \\
& MEGU                      
& 92.27 ± 0.06 & 92.08 ± 0.06     & 80.13 ± 0.01    & 77.18 ± 4.47       \\
& FedLU
&91.16 ± 0.06 &89.84 ± 1.03 &\textbf{83.26 ± 0.37} &84.74 ± 0.46 \\
& FedDM
&83.36 ± 2.89 &91.81 ± 0.30 &76.63 ± 1.13 &72.49 ± 3.59 \\
& PAGE
&\textbf{92.66 ± 0.08} &\textbf{92.76 ± 0.06} &82.04 ± 0.33 &\textbf{82.76 ± 0.59} \\
\midrule[0.3pt]
\end{tabular}
}
}

\label{tab: unlearn_set}
\end{table}
In this section, we conduct a comprehensive experimental evaluation of PAGE. First, we systematically introduce the datasets used in the experiments. Second, we present state-of-the-art(SOTA) methods for FU and GU employed for comparison. Finally, we describe the specific unlearning configurations and evaluation methodologies. Overall, we aim to address the following research questions:
\textbf{Q1}: Compared with the SOTA federated unlearning and graph unlearning methods, can PAGE achieve the best performance under multiple unlearn requests?
\textbf{Q2}: How does PAGE achieve the dual advantages of maintaining model utility and forgetting integrity?
\textbf{Q3}: Does PAGE demonstrates strong robustness across diverse unlearning scenarios?

\subsection{Experimental Setup}
All experiments are conducted on a system equipped with an NVIDIA A100 80GB PCIe GPU and an Intel(R) Xeon(R) Gold 6240 CPU @ 2.60GHz, with CUDA Version 12.4 enabled. The software environment is set up with Python 3.9.16 and PyTorch 1.13.1 to ensure optimal compatibility and performance for all algorithms.

\textbf{Datasets}.
To comprehensively evaluate various unlearning methods, we collect real-world datasets of varying sizes from various domains (e.g., citation networks, co-author networks, etc.). 
From small-scale dataset Cora~\cite{Yang16cora} to medium-scale dataset Photo~\cite{shchur2018amazon_datasets} to large-scale dataset ogbn-arxiv~\cite{hu2020ogb}, we assign different numbers of clients according to the size of the dataset.
Detailed data partitioning and simulation strategies refer to Appendix.

\textbf{Baselines}.
We list Retrain and compare PAGE with 10 baseline methods.
Most existing methods do not simultaneously support meta unlearning and client unlearning. Consequently, we divide the comparative experiments into two sets:
(1) Client unlearning: FedEraser~\cite{romandini2024federaser}, FUSED~\cite{zhong2025fused}, MoDe~\cite{zhao2024mode}, ReGEnUnlearn~\cite{liu2025subgraph}.
(2) Meta unlearning: GIF~\cite{wu2023gif}, CEU~\cite{2023WuCEU}, D2DGN~\cite{2024D2DGN}, MEGU~\cite{xkliMEGU2024}, FedLU~\cite{zhu2023fedlu}, FedDM~\cite{liu2024feddm}.

\textbf{Unlearning Settings}.
For the three unlearning modes in meta-unlearning, we randomly unlearn $10\%$ of relevant graph data. Specifically, we randomly unlearn $10\%$ of nodes for node unlearn request, $10\%$ of edges for edge unlearn request and $10\%$ of node features for feature unlearn request. For client unlearn, $20\%$ of clients are randomly selected as unlearning clients, where all local data need to be unlearned.

\textbf{Metrics}. To evaluate the prediction performance of the model after unlearning, we use accuracy as the performance metric. For meta unlearning, we calculate the prediction accuracy of the local model on the local dataset. For client unlearn, we calculate the prediction accuracy of the global model on the local dataset. To rigorously validate unlearning effects, we employ membership inference attack~\cite{olatunji2021mia}(MIA), which can further determine whether the target instance has been effectively removed from the training set underlying the unlearned model.

\begin{figure}[t]
  \includegraphics[width=\linewidth]{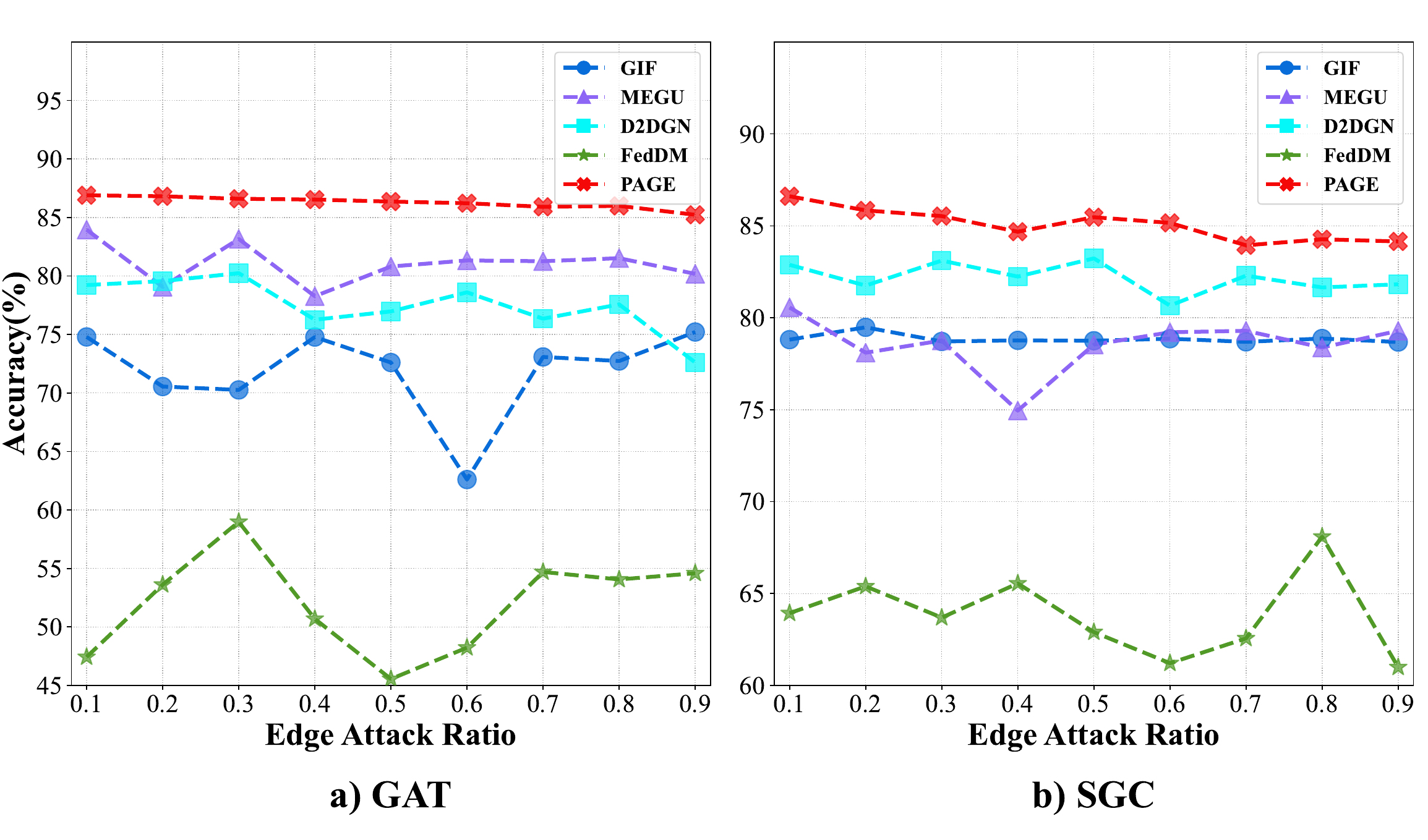}
  \caption{Performance on Photo under edge attack.}
  \label{fig: Edge Attack}
\end{figure}

\subsection{Q1: Performance Comparison}
To answer Q1, we show the prediction performance comparison of each method under different unlearn request settings in Table \ref{Q1}. The results show that PAGE can outperform the baseline level on most datasets. On average, PAGE exhibits a remarkable average improvement of 5.08\% over the SOTA approach under client unlearn request and 1.50\% under meta unlearn request.
It is noteworthy that on relatively small-scale datasets, the performance gap between PAGE and other methods is not statistically significant. However, on larger-scale datasets, PAGE demonstrates substantial performance advantages. For instance, under client unlearn requests, PAGE achieves merely a 2.13\% improvement over SOTA methods on Photo, whereas it attains a 11.84\% performance gain over SOTA methods on ogbn-arxiv.
Additionally, for the other two granularity levels of unlearn requests (edge/feature unlearn) within meta-unlearn, we further evaluate PAGE's performance against comparative methods.
As shown in Table \ref{tab: unlearn_set}, we compared the prediction performance of PAGE and other methods on Physics and Computers under simulation strategy metis-based community split (Metis) and metis-based label imbalance split (Metis-Plus). Except for some combinations, the prediction performance of PAGE is not as good as FedLU, but it is significantly ahead of other methods in other combinations.

\begin{figure*}[t]
  \includegraphics[width=\textwidth]{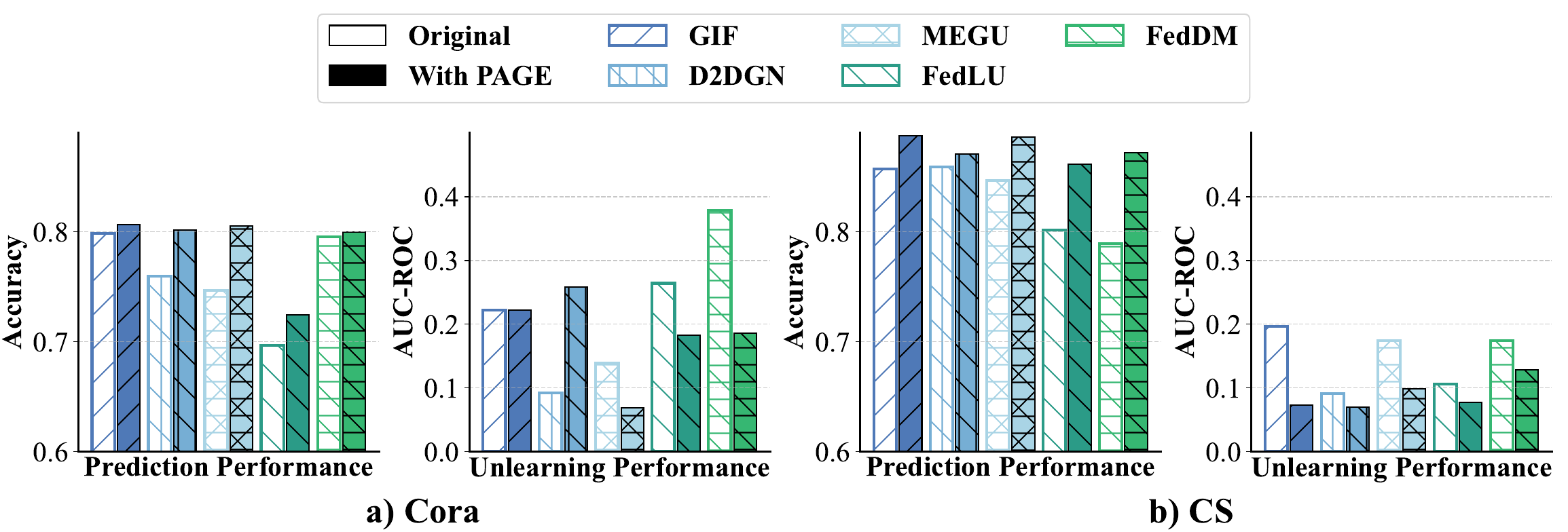}
  \caption{
    Augmentation study on GU methods. The left column represents original, the right represents combined with PAGE.}
  \label{fig: Ablation}
\end{figure*}

\begin{figure}[t]
  \includegraphics[width=\linewidth]{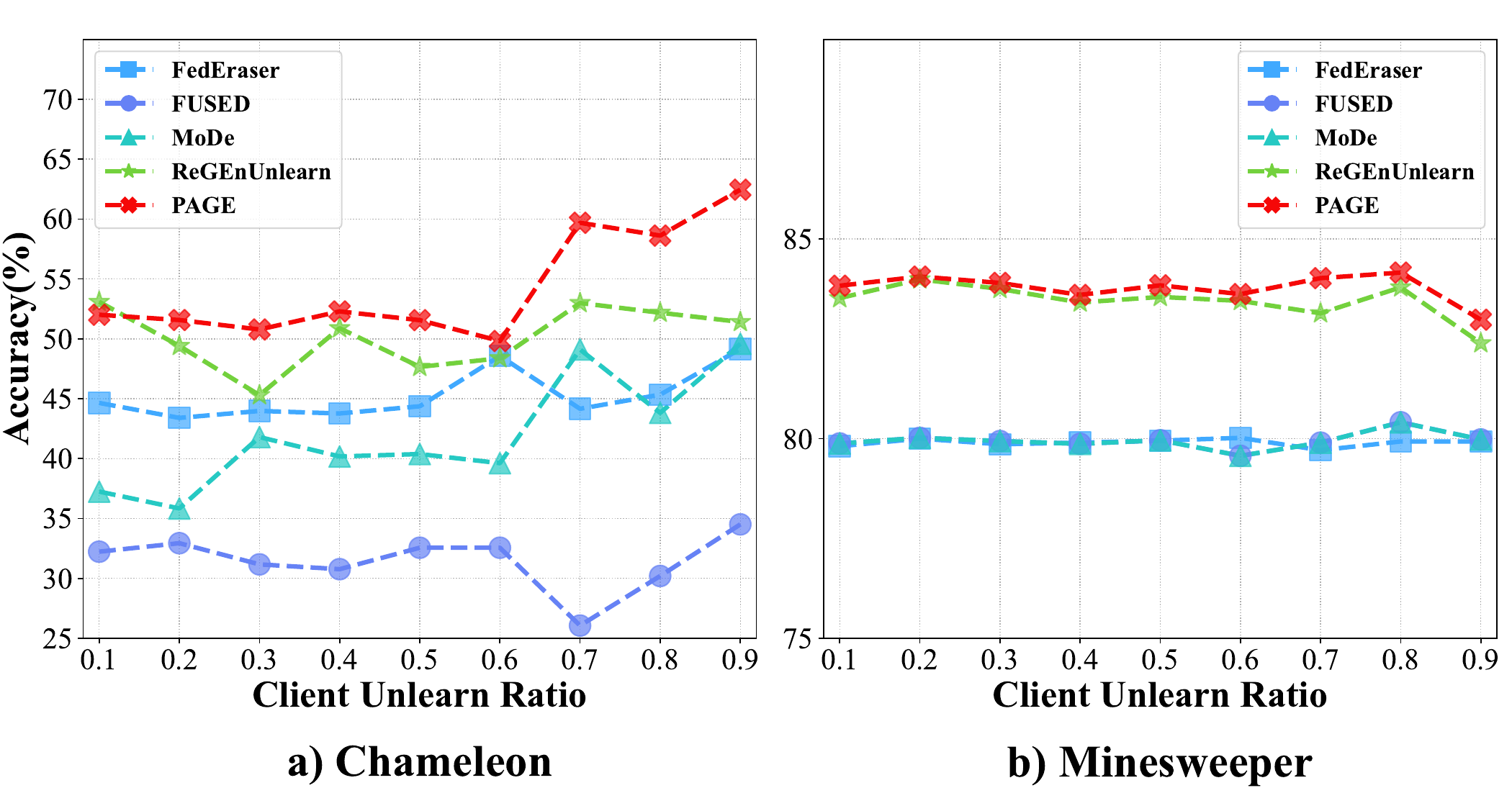}
  \caption{
    Performance under differnet client unlearn ratio.}
  \label{fig: unlearn_ratio}
\end{figure}

To answer Q1 from the perspective of unlearning capability, Fig.~\ref{fig: Edge Attack} presents the unlearning capabilities of various methods under varying intensities of edge attack settings under meta unlearning.
The edge-attack configuration involves injecting poisoned edges into each client's data before federated learning to degrade original model performance.
After executing FGU algorithms, the models should regain their original performance, where the unlearning efficacy is quantified by the prediction accuracy (higher accuracy indicates stronger unlearning capability).
This experiment employs the GAT~\cite{velivckovic2017gat} and SGC~\cite{wu2019sgc} on Photo configurations to evaluate the prediction performance between methods under different edge-attack ratios.
As demonstrated in Fig.~\ref{fig: Edge Attack}, PAGE consistently maintains SOTA accuracy under multiple attack ratios, exhibiting minimal sensitivity to increasing attack intensity. For instance, on Photo, PAGE achieves 2.94\% to 8.62\% higher accuracy than comparative SOTA methods. These results substantiate that PAGE delivers superior unlearning capabilities.

\subsection{Q2: Augmentation Study}
In order to systematically verify the effectiveness of the components of our proposed method, we designed a series of exhaustive augmentation experiments to answer Q2.
A key point of our method design is the dual role of the influenced unlearn component: in the client unlearn scenario, it operates independently as a core processing engine; while in the meta unlearn scenario, it acts as a plug-and-play enhancement plugin.
The performance comparison in Q1 has verified the effectiveness of this component in client unlearn scenario, and here we verify its effectiveness as an enhancement plugin in meta unlearn scenario.
As shown in Fig.~\ref{fig: Ablation}, in the meta unlearn scenario, we evaluate the prediction performance and unlearning performance of the existing graph unlearning method before and after combining with the influenced unlearn component. As can be seen in the figure, the prediction performance and unlearning performance of the existing local unlearn method are significantly improved after combining with the influence module.
Specifically, on Cora, the influenced unlearn component yields an average improvement of 2.82\% in prediction performance and 4.64\% in unlearning performance compared to the base method. On CS, it further enhances prediction performance by 4.49\% and unlearning performance by 7.22\%.

\subsection{Q3: Robustness Analysis}

To answer Q3, we designed experiments that evaluate model performance under varying unlearn intensities. Using GraphSAGE as the backbone under client unlearn with louvain simulation, we measured prediction accuracy across different unlearn ratios on Chameleon and Minesweeper.

As shown in Fig.~\ref{fig: unlearn_ratio}, PAGE maintains superior performance over counterparts despite increasing unlearn intensity. On dataset Chameleon, while slightly underperforming RFPS at low unlearn intensities, PAGE demonstrates minimal fluctuation and remains stable as intensity escalates. Notably, at high forgetting intensities, PAGE exhibits counterintuitive performance gains—increasing accuracy over baselines and widening the performance gap. From the above experiments, it can be seen that compared with other methods, PAGE has stronger robustness.

Furthermore, Fig.~\ref{fig: Edge Attack} reveals that as the edge-attack ratio progressively increases, PAGE persistently maintains prediction stability without exhibiting significant performance fluctuations observed in other methods. This robustness against adversarial edge perturbations substantiates PAGE's exceptional resilience.

\section{Conclusion}

This paper critically analyzes federated and graph unlearning methods, revealing incomplete unlearning support and residual knowledge permeation as fundamental limitations. To address these gaps, we introduce the first multi-scenario federated graph unlearning framework. Our approach pioneers prototype matching while integrating negative knowledge distillation to systematically eliminate knowledge permeation.
Core innovations include:
(1) Utilizing prototype vectors as knowledge carriers for precise local unlearning;
(2) Extracting unlearned knowledge via adversarial graph samples;
(3) Directing cross-client influence unlearning to eradicate permeation impacts without privacy compromise.
Future research must prioritize integrated frameworks harmonizing local and influence unlearning for comprehensive permeation mitigation.

\bibliography{aaai2026}

\begin{thebibliography}{31}
\providecommand{\natexlab}[1]{#1}

\bibitem[{Cai et~al.(2023)Cai, Huang, Xia, and Ren}]{cai2023app_gnn_rec3}
Cai, X.; Huang, C.; Xia, L.; and Ren, X. 2023.
\newblock LightGCL: Simple Yet Effective Graph Contrastive Learning for Recommendation.
\newblock In \emph{International Conference on Learning Representations, ICLR}.

\bibitem[{Chang and Shokri(2023)}]{chang2023bias}
Chang, H.; and Shokri, R. 2023.
\newblock Bias propagation in federated learning.
\newblock \emph{arXiv preprint arXiv:2309.02160}.

\bibitem[{Dong et~al.(2024)Dong, Zhang, Lei, Zou, and Li}]{2024IDEA}
Dong, Y.; Zhang, B.; Lei, Z.; Zou, N.; and Li, J. 2024.
\newblock IDEA: A Flexible Framework of Certified Unlearning for Graph Neural Networks.
\newblock In \emph{Proceedings of the 30th ACM SIGKDD Conference on Knowledge Discovery and Data Mining}, 621–630. Association for Computing Machinery.
\newblock ISBN 9798400704901.

\bibitem[{Hamilton, Ying, and Leskovec(2017)}]{hamilton2017graphsage}
Hamilton, W.; Ying, Z.; and Leskovec, J. 2017.
\newblock Inductive representation learning on large graphs.
\newblock \emph{Advances in Neural Information Processing Systems, NeurIPS}.

\bibitem[{He et~al.(2021)He, Balasubramanian, Ceyani, Yang, Xie, Sun, He, Yang, Yu, Rong et~al.}]{he2021fedgraphnn}
He, C.; Balasubramanian, K.; Ceyani, E.; Yang, C.; Xie, H.; Sun, L.; He, L.; Yang, L.; Yu, P.~S.; Rong, Y.; et~al. 2021.
\newblock Fedgraphnn: A federated learning system and benchmark for graph neural networks.
\newblock \emph{arXiv preprint arXiv:2104.07145}.

\bibitem[{Hu et~al.(2020)Hu, Fey, Zitnik, Dong, Ren, Liu, Catasta, and Leskovec}]{hu2020ogb}
Hu, W.; Fey, M.; Zitnik, M.; Dong, Y.; Ren, H.; Liu, B.; Catasta, M.; and Leskovec, J. 2020.
\newblock Open graph benchmark: Datasets for machine learning on graphs.
\newblock \emph{Advances in Neural Information Processing Systems, NeurIPS}.

\bibitem[{Kipf and Welling(2017)}]{kipf2016gcn}
Kipf, T.~N.; and Welling, M. 2017.
\newblock Semi-supervised classification with graph convolutional networks.
\newblock In \emph{International Conference on Learning Representations, ICLR}.

\bibitem[{Li et~al.(2024)Li, Zhao, Wu, Zhang, Li, and Wang}]{xkliMEGU2024}
Li, X.; Zhao, Y.; Wu, Z.; Zhang, W.; Li, R.-H.; and Wang, G. 2024.
\newblock Towards Effective and General Graph Unlearning via Mutual Evolution.
\newblock \emph{Proceedings of the AAAI Conference on Artificial Intelligence}, 38(12): 13682--13690.

\bibitem[{Liu and Fang(2024)}]{liu2024feddm}
Liu, B.; and Fang, Y. 2024.
\newblock Federated knowledge graph unlearning via diffusion model.
\newblock \emph{arXiv preprint arXiv:2403.08554}.

\bibitem[{Liu and Liu(2025)}]{liu2025subgraph}
Liu, F.; and Liu, H. 2025.
\newblock Subgraph Federated Unlearning.
\newblock In \emph{THE WEB CONFERENCE 2025}.

\bibitem[{Olatunji, Nejdl, and Khosla(2021)}]{olatunji2021mia}
Olatunji, I.~E.; Nejdl, W.; and Khosla, M. 2021.
\newblock Membership Inference Attack on Graph Neural Networks.
\newblock In \emph{2021 Third IEEE International Conference on Trust, Privacy and Security in Intelligent Systems and Applications (TPS-ISA)}, 11--20.

\bibitem[{Pardau(2018)}]{CCPA1}
Pardau, S.~L. 2018.
\newblock The california consumer privacy act: Towards a european-style privacy regime in the united states.
\newblock \emph{J. Tech. L. \& Pol'y}, 23: 68.

\bibitem[{Platonov et~al.(2023)Platonov, Kuznedelev, Diskin, Babenko, and Prokhorenkova}]{platonov2023hete_gnn_survey4}
Platonov, O.; Kuznedelev, D.; Diskin, M.; Babenko, A.; and Prokhorenkova, L. 2023.
\newblock A critical look at the evaluation of GNNs under heterophily: are we really making progress?
\newblock \emph{International Conference on Learning Representations, ICLR}.

\bibitem[{Qu et~al.(2023)Qu, Yao, Liu, and Wang}]{qu2023app_gnn_bio2}
Qu, Z.; Yao, T.; Liu, X.; and Wang, G. 2023.
\newblock A Graph Convolutional Network Based on Univariate Neurodegeneration Biomarker for Alzheimer’s Disease Diagnosis.
\newblock \emph{IEEE Journal of Translational Engineering in Health and Medicine}.

\bibitem[{Regulation(2018)}]{GDPR}
Regulation, G. D.~P. 2018.
\newblock General data protection regulation (GDPR). ntersoft Consulting.
\newblock \emph{Obtenido de https://www. epsu. org/sites/> default/files/article/files/GDPR\_FINAL\_EPSU. pdf}.

\bibitem[{Romandini et~al.(2024)Romandini, Mora, Mazzocca, Montanari, and Bellavista}]{romandini2024federaser}
Romandini, N.; Mora, A.; Mazzocca, C.; Montanari, R.; and Bellavista, P. 2024.
\newblock Federated unlearning: A survey on methods, design guidelines, and evaluation metrics.
\newblock \emph{IEEE Transactions on Neural Networks and Learning Systems}.

\bibitem[{Rozemberczki, Allen, and Sarkar(2021)}]{Rozemberczkichameleon}
Rozemberczki, B.; Allen, C.; and Sarkar, R. 2021.
\newblock {Multi-Scale attributed node embedding}.
\newblock \emph{Journal of Complex Networks}, 9(2).

\bibitem[{Shchur et~al.(2018)Shchur, Mumme, Bojchevski, and G{\"u}nnemann}]{shchur2018amazon_datasets}
Shchur, O.; Mumme, M.; Bojchevski, A.; and G{\"u}nnemann, S. 2018.
\newblock Pitfalls of graph neural network evaluation.
\newblock \emph{arXiv preprint arXiv:1811.05868}.

\bibitem[{Sinha, Mandal, and Kankanhalli(2024)}]{2024D2DGN}
Sinha, Y.; Mandal, M.; and Kankanhalli, M. 2024.
\newblock Distill to Delete: Unlearning in Graph Networks with Knowledge Distillation.
\newblock arXiv:2309.16173.

\bibitem[{Tolpegin et~al.(2020)Tolpegin, Truex, Gursoy, and Liu}]{tolpegin2020data}
Tolpegin, V.; Truex, S.; Gursoy, M.~E.; and Liu, L. 2020.
\newblock Data poisoning attacks against federated learning systems.
\newblock In \emph{European symposium on research in computer security}, 480--501. Springer.

\bibitem[{Veli{\v{c}}kovi{\'c} et~al.(2018)Veli{\v{c}}kovi{\'c}, Cucurull, Casanova, Romero, Lio, and Bengio}]{velivckovic2017gat}
Veli{\v{c}}kovi{\'c}, P.; Cucurull, G.; Casanova, A.; Romero, A.; Lio, P.; and Bengio, Y. 2018.
\newblock Graph attention networks.
\newblock In \emph{International Conference on Learning Representations, ICLR}.

\bibitem[{Wang et~al.(2018)Wang, Chen, Ren, Yu, Cheng, and Lin}]{ijcai2018p142}
Wang, Z.; Chen, T.; Ren, J.; Yu, W.; Cheng, H.; and Lin, L. 2018.
\newblock Deep Reasoning with Knowledge Graph for Social Relationship Understanding.
\newblock In \emph{Proceedings of the Twenty-Seventh International Joint Conference on Artificial Intelligence, {IJCAI-18}}, 1021--1028. International Joint Conferences on Artificial Intelligence Organization.

\bibitem[{Wu et~al.(2019)Wu, Souza, Zhang, Fifty, Yu, and Weinberger}]{wu2019sgc}
Wu, F.; Souza, A.; Zhang, T.; Fifty, C.; Yu, T.; and Weinberger, K. 2019.
\newblock Simplifying graph convolutional networks.
\newblock In \emph{International Conference on Machine Learning, ICML}.

\bibitem[{Wu et~al.(2023{\natexlab{a}})Wu, Yang, Qian, Sui, Wang, and He}]{wu2023gif}
Wu, J.; Yang, Y.; Qian, Y.; Sui, Y.; Wang, X.; and He, X. 2023{\natexlab{a}}.
\newblock GIF: A General Graph Unlearning Strategy via Influence Function.
\newblock In \emph{Proceedings of the ACM Web Conference, WWW}.

\bibitem[{Wu et~al.(2023{\natexlab{b}})Wu, Shen, Ning, Wang, and Wang}]{2023WuCEU}
Wu, K.; Shen, J.; Ning, Y.; Wang, T.; and Wang, W.~H. 2023{\natexlab{b}}.
\newblock Certified Edge Unlearning for Graph Neural Networks.
\newblock In \emph{Proceedings of the 29th ACM SIGKDD Conference on Knowledge Discovery and Data Mining}, 2606–2617. Association for Computing Machinery.
\newblock ISBN 9798400701030.

\bibitem[{Wu et~al.(2020)Wu, Pan, Chen, Long, Zhang, and Yu}]{wu2020comprehensive}
Wu, Z.; Pan, S.; Chen, F.; Long, G.; Zhang, C.; and Yu, P.~S. 2020.
\newblock A comprehensive survey on graph neural networks.
\newblock \emph{IEEE transactions on neural networks and learning systems}, 32(1): 4--24.

\bibitem[{Yang et~al.(2024)Yang, Han, Chai, Ebrahimi, Behnia, and Padmanabhan}]{yang2024machine}
Yang, Y.; Han, X.; Chai, Y.; Ebrahimi, R.; Behnia, R.; and Padmanabhan, B. 2024.
\newblock From Machine Learning to Machine Unlearning: Complying with GDPR's Right to be Forgotten while Maintaining Business Value of Predictive Models.
\newblock \emph{arXiv preprint arXiv:2411.17126}.

\bibitem[{Yang, Cohen, and Salakhutdinov(2016)}]{Yang16cora}
Yang, Z.; Cohen, W.~W.; and Salakhutdinov, R. 2016.
\newblock Revisiting Semi-Supervised Learning with Graph Embeddings.
\newblock In \emph{International Conference on Machine Learning, ICML}.

\bibitem[{Zhao et~al.(2024)Zhao, Wang, Qi, Huang, Wei, and Zhang}]{zhao2024mode}
Zhao, Y.; Wang, P.; Qi, H.; Huang, J.; Wei, Z.; and Zhang, Q. 2024.
\newblock Federated Unlearning With Momentum Degradation.
\newblock \emph{IEEE Internet of Things Journal}, 11(5): 8860--8870.

\bibitem[{Zhong et~al.(2025)Zhong, Bao, Wang, Zhang, Zhou, Lyu, and Lim}]{zhong2025fused}
Zhong, Z.; Bao, W.; Wang, J.; Zhang, S.; Zhou, J.; Lyu, L.; and Lim, W. Y.~B. 2025.
\newblock Unlearning through knowledge overwriting: Reversible federated unlearning via selective sparse adapter.
\newblock In \emph{Proceedings of the Computer Vision and Pattern Recognition Conference}, 30661--30670.

\bibitem[{Zhu, Li, and Hu(2023)}]{zhu2023fedlu}
Zhu, X.; Li, G.; and Hu, W. 2023.
\newblock Heterogeneous federated knowledge graph embedding learning and unlearning.
\newblock In \emph{Proceedings of the ACM web conference 2023}, 2444--2454.

\end{thebibliography}

\clearpage
\appendix

\section{Appendix}

\subsection{Datasets}
The following describes the dataset information used in all experiments. For the specific data and dataset division, please refer to Table \ref{tab: datasets}.

\begin{table*}[t]
    \caption{The statistics of the experimental datasets.
    }
    \label{tab: datasets}
    \resizebox{\linewidth}{32mm}{
    \setlength{\tabcolsep}{1.5mm}{
    \begin{tabular}{ccccccc}
    \midrule[0.3pt]
    Dataset          & Nodes & Features & Edges    & Classes & Train/Val/Test        & Description         \\ \midrule[0.3pt]
    Cora             & 2,708   & 1,433      & 5,429      & 7    & 20\%/40\%/40\%  & citation network    \\
    PubMed           & 19,717  & 500        & 44,338     & 3         & 20\%/40\%/40\%             & citation network    \\
    ogbn-arxiv & 169343 & 128 & 231559 & 40 & 60\%/20\%/20\% & citation network    \\
    \midrule[0.3pt]
    Amazon Photo     & 7,487   & 745        & 119,043    & 8         & 20\%/40\%/40\%                 & co-purchase graph \\ 
    Amazon Computers  & 13,381  & 767        & 245,778    & 10        & 20\%/40\%/40\%               & co-purchase graph \\ \midrule[0.3pt]
    Coauthor CS      & 18,333  & 6,805      & 81,894     & 15        & 20\%/40\%/40\%             & co-authorship graph \\
    Coauthor Physics & 34,493  & 8,415      & 247,962    & 5         & 20\%/40\%/40\%            & co-authorship graph \\ 
    \midrule[0.3pt]
    Tolokers              & 11,758  & 10         & 519,000    & 2       & 50\%/25\%/25\%               & crowd-sourcing network     \\
    Minesweeper           & 10,000  & 7        & 39,402    & 2         & 50\%/25\%/25\%               & game synthetic network       \\ 
    Amazon-ratings           & 24,292  & 300        & 93,050    & 5         & 50\%/25\%/25\%               & rating network       \\
    Chameleon           & 2,277  & 2,325        & 36,101    & 5         & 48\%/32\%/20\%               & rating network       \\
    \midrule[0.3pt]
    \end{tabular}
    }}
\end{table*}

\textbf{Cora} and \textbf{PubMed}~\cite{Yang16cora} are widely used citation network datasets in the field of federated graph unlearning. In these datasets, nodes represent research papers, and edges denote citation relationships between papers. Each node is characterized by word vectors, with each being uniquely associated with a specific category label, indicating the presence or absence of specific words in each paper. These datasets are frequently employed in tasks such as node classification, providing a reliable basis for evaluating model performance.

\textbf{CS} and \textbf{Physics}~\cite{shchur2018amazon_datasets}, as co-authorship graphs originating from Microsoft Academic Graph, are specially designed for node classification tasks. In these datasets, academic collaboration is depicted such that nodes stand for authors, and an edge emerges between two nodes when the corresponding authors have co-written a paper. For both datasets, node features are denoted by keywords related to the papers each author has published, and class labels mark the research fields where each author is most active, offering abundant semantic information for classification tasks.

\textbf{Photo} and \textbf{Computers}~\cite{shchur2018amazon_datasets} datasets, originating from Amazon's co-purchase graph, are designed to illustrate the relationships between products that are frequently purchased together. Within these datasets, nodes represent single products, and edges signify how often consumers buy these products in combination—thus mirroring the market trends and consumer behaviors on Amazon's e-commerce platform. The Photo dataset is dedicated to photographic gear, while the Computers dataset focuses on computer-related products; both are commonly employed in tasks like node classification.

\textbf{Tolokers}~\cite{platonov2023hete_gnn_survey4} dataset is sourced from a crowdsourcing platform, with the goal of predicting which workers have been banned from one of the projects. Nodes in the dataset represent workers who have participated in at least one of the 13 selected projects, and an edge links two workers if they have collaborated on the same task.

\textbf{Minesweeper}~\cite{platonov2023hete_gnn_survey4} dataset, drawn from an online gaming platform, simulates interactions within the Minesweeper gaming environment. In this graph, each node stands for a player, and edges represent instances of collaboration or competition between players during gameplay. Node attributes are derived from player stats and in-game metrics, offering a thorough overview of user behavior patterns. This dataset holds significance for examining social connection and behavior trends, and proves useful in evaluating unlearning strategies within dynamic, interaction-based networks.

\textbf{Amazon-ratings}~\cite{platonov2023hete_gnn_survey4} dataset documents user interactions with products on Amazon. Within this dataset, nodes represent products, and edges link items that consumers frequently purchase together. Node features are built on FastText embeddings of words from product descriptions. The dataset’s task is to predict the average ratings given by reviewers, and it plays a significant role in tasks like forecasting user preferences and understanding product correlations. It also serves as a real-world testing ground for applying graph neural networks in large-scale scenarios.

\textbf{ogbn-arxiv}~\cite{hu2020ogb} is a comprehensive academic graph from Microsoft Academic Graph (MAG), created to support machine learning tasks on graph data. It presents a citation network of arXiv papers, where each paper is represented by the average word embeddings from its title and abstract. The graph structure of ogbn-arxiv features scientific papers as nodes and citation links between papers as edges, clearly reflecting academic referencing connections. It plays an important role in tasks such as node classification.

\textbf{Chameleon}~\cite{Rozemberczkichameleon} is a page-to-page network extracted from specific topics on Wikipedia. In the dataset, nodes stand for web pages, and edges indicate mutual links between them. Node features are derived from several informative nouns on Wikipedia, obtained through text processing techniques. The dataset classifies nodes based on the average monthly traffic of the web pages. Its primary task is to classify nodes using their features and the graph structure. Due to its complex graph structure and rich node features, it is often used to test the performance of graph neural networks.

\subsection{GNN Backbone}
\textbf{GCN}~\cite{kipf2016gcn} proposes a novel method based on a first-order approximation of spectral convolution on graphs. By aggregating the features of neighboring nodes, it learns hidden layer representations, which can effectively encode local graph structures and node features, reduce computational complexity, and is suitable for tasks such as node classification, promoting the development of deep learning on graph data.

\textbf{GAT}~\cite{velivckovic2017gat} employs attention mechanisms to measure how significant each neighbor is to message aggregation, which in turn boosts the model's representational power and generalization ability. This approach allows for the implicit allocation of varying weights to different nodes in the neighborhood, free from any prior dependence on graph structures. It enhances adaptability to node relationships and is suitable for a range of graph learning tasks.

\textbf{GraphSAGE}~\cite{hamilton2017graphsage} leverages attribute information from neighboring nodes to efficiently generate representations and is specifically designed to address the challenges of large-scale graph data. This method features a general inductive framework that utilizes sampled and aggregated node feature information to efficiently generate node embeddings for unseen new nodes.

\textbf{SGC}~\cite{wu2019sgc} simplifies GCN by stripping out non-linearities and collapsing weight matrices between consecutive layers, thus forming a linear model. Theoretical analyses indicate that this simplified model is equivalent to a fixed low-pass filter followed by a linear classifier, boasting fewer parameters and greater efficiency.

\subsection{Baseline}
\textbf{FedEraser}~\cite{romandini2024federaser}, as the first federated unlearning method, addresses the challenge of efficiently removing data from federated learning models. It reconstructs the model by leveraging historical parameters retained by the central server and puts forward a novel calibration method for updating the retained parameters. The effectiveness and efficiency exhibited by this method hold significant importance in the field of federated learning.

\textbf{FUSED}~\cite{zhong2025fused} innovatively achieves federated unlearning through knowledge overwriting. Its core lies in first analyzing the sensitivity of each model layer to knowledge changes via Critical Layer Identification (CLI), locating sensitive layers and constructing sparse unlearning adapters for them. Then, the original model parameters are frozen, and only the adapters are trained to overwrite the knowledge to be forgotten with remaining knowledge. This method enables reversible unlearning through independent adapters, significantly reduces computational and communication costs of unlearning, and achieves unlearning performance comparable to retraining, outperforming other baselines remarkably while effectively alleviating the problem of indiscriminate unlearning.

\textbf{MoDe}~\cite{zhao2024mode} effectively improves efficiency and accuracy on remaining data nodes by decomposing the unlearning process into two steps: knowledge erasure and memory guidance. It innovatively applies momentum degradation to the knowledge erasure strategy to erase implicit knowledge in the model. Moreover, this training-independent and effective method can handle unlearning requests from clients and categories.

\textbf{ReGEnUnlearn}~\cite{liu2025subgraph} proposes a subgraph federated unlearning framework to address the unique structural dependency and cross-client interference issues of graph data. Through RFPS learning optimal sampling strategies to reduce cross-client subgraph interference, combined with the Parameter-free Graph Prompt Knowledge Distillation (PGPKD) module to extract the unique knowledge of target clients, it achieves comprehensive unlearning of multiple clients, demonstrating high unlearning performance and efficiency.

\textbf{GIF}~\cite{wu2023gif} innovatively addresses graph unlearning (GU) by incorporating an influence function for affected domains, enhancing the efficiency and accuracy of the unlearning process. It redefines influence by adding a loss term for affected neighborhoods to handle dependencies between adjacent nodes, and further derives a closed-form solution for parameter changes as a basis. Using a universal formula that unifies different types of unlearning tasks, GIF achieves significant advantages in terms of unlearning efficiency and performance in the field of graph unlearning.

\textbf{D2DGN}~\cite{2024D2DGN} skillfully addresses the challenge of deleting specific elements in graph unlearning through knowledge distillation. By dividing and marking the complete graph knowledge, and performing distillation using response-based soft targets and feature-based node embeddings, it achieves unlearning results with higher efficiency, better performance, and no additional overhead costs. D2DGN effectively eliminates the influence of deleted elements while preserving the required knowledge, playing a significant role in the field of graph unlearning.

\textbf{MEGU}~\cite{xkliMEGU2024} introduces a groundbreaking mutual evolution approach for graph unlearning (GU), where the predictive module and the unlearning module co-evolve within a single training framework, and efficiency is enhanced through Adaptive HIN Selection. The adaptability of MEGU enables it to demonstrate favorable performance and training efficiency in meeting unlearning requirements at the feature, node, and edge levels, showing strong advantages in the field of GU.

\textbf{FedLU}~\cite{zhu2023fedlu} proposes a framework for heterogeneous federated knowledge graph embedding learning and unlearning. To address the problems of data heterogeneity and knowledge forgetting in federated knowledge graph embedding, this method adopts knowledge distillation in the learning phase to mitigate the impact of heterogeneity; in the unlearning phase, it combines cognitive neuroscience to eliminate specific triple knowledge and propagates the unlearning effect to the global model through knowledge distillation. This method demonstrates good unlearning performance and high unlearning efficiency. 

\textbf{FedDM}~\cite{liu2024feddm} is a new framework specifically designed for the federated training of diffusion models. Through rigorous theoretical analysis, it proves the convergence of the proposed model and the corresponding conditions. The experimental results show high communication efficiency and model convergence, and can maintain high generation quality at different resolutions in terms of image modality.

\subsection{Federated Learning Simulation Mode}

\textbf{Metis-based Community Split} is a widely adopted federated data simulation strategy in SubgraphFL, enabling efficient data partitioning and effective sharing of graph structure knowledge. This method simplifies the graph through edge weight matching, performs K-way partitioning, and iteratively divides the global graph into tightly connected communities by leveraging hierarchical graph partitioning capabilities.

\textbf{Louvain-based Community Split} is another popular federated data simulation strategy in Subgraph-FL, whose core lies in partitioning the global graph using the Louvain algorithm. It identifies community structures by maximizing the modularity score, rapidly accomplishes the identification of communities in the global graph and provides the basis for partitioning, and finally performs partitioning according to actual needs.

\textbf{Metis-based Label Imbalance Split} is a Subgraph-FL data simulation strategy based on label distribution similarity. To address the challenge in controlling subgraph heterogeneity among clients caused by the lack of post-processing capabilities in the naive Metis-based Community Split, this strategy first defines predefined community partitions, then performs clustering based on label distribution similarity, thereby merging similar communities under a single client.

\textbf{Louvain-based Label Imbalance Split} builds on the traditional Louvain method by allocating communities to clients in line with the similarity of their label distributions. This approach ensures that each client receives communities with consistent label features, thereby easing label imbalance and enabling fair model training across federated clients. By adjusting label distributions, the strategy enhances the fairness and robustness of federated learning, reducing potential biases caused by heterogeneous label distributions among clients.

\subsection{Hyperparameter Settings}
To help reproduce our work, we give the hyperparameters related to the training process on each dataset in Table \ref{tab: hyperparameters}.

\subsection{Experimental Details}

In this part we will introduce the standardized experimental protocol of the experiments mentioned in the main text. 

\textbf{Performance Comparison}. This experiment uses $8$ benchmark datasets and distribute the dataset to different numbers of clients based on the size of the dataset. Specifically, distribute Cora and PubMed to $5$ clients; distribute CS and Photo to $10$ clients; distribute Tolokers and Minesweeper to $15$ clients; distribute Amazon-ratings and ogbn-arxiv to $20$ clients. The simulation mode is louvain-based community split and the backbone is selected as GCN.

In Table \ref{Q1}, for client unlearning, the unlearn ratio is set to $0.2$, and $0.1$ for meta unlearning (node unlearning).
In Table \ref{tab: unlearn_set}, on Computers and Physics, we employ metis and metis-plus as simulation mode. The unlearn ratio is set to $0.1$ and the backbone is selected as GAT.
In experiment of Fig.\ref{fig: Edge Attack}, we choose two backbone: GAT and SGC on Photo. And the simulation mode is louvain-based community split. As the figure show, the unlearn ratio increases from $0.1$ to $0.9$. 

\textbf{Augmentation Study}
In the experiment of augmentation study, we choose SGC as the backbone and louvain-based community split as the simulation mode under meta unlearning (node unlearning) on Cora and CS. The unlearn ratio is set to 0.1.

\textbf{Robustness Analysis}
In the experiment of robustness analysis, we choose GraphSage as the backbone and louvain-based community split as the simulation mode under client unlearn on Chameleon and Minesweeper. The unlearn ratio increases from 0.1 to 0.9.


\begin{table}[t]
\caption{Detailed hyperparameter setting on all datasets.}
\centering
\label{tab: hyperparameters}
\resizebox{0.9\linewidth}{25mm}{
\setlength{\tabcolsep}{1.2mm}{
\begin{tabular}{cccccc}
\midrule[0.3pt]
Dataset          & lr & \makecell{train\\epoches}   & \makecell{client unlearn \\epoches} & \makecell{meta unlearn \\epoches}& $\alpha$  \\ \midrule[0.3pt]
Cora   & 1e-3 & 10 & 2 & 1 & 5.0   \\ 
PubMed  & 1e-3 & 10 & 5 & 5 & 1.0  \\ 
obgn-arxiv & 5e-3 & 50 & 25 & 50 & 1.0 \\
\midrule[0.3pt]
Amazon Photo & 1e-3 & 10 & 5 & 2  & 1.0   \\ 
Amazon Computers & 1e-3 & 10 & 5 & 10 & 1.0\\ 
\midrule[0.3pt]
Coauthor CS  & 1e-3 & 10 & 5 & 5 & 1.0  \\
Coauthor Physics & 1e-3 & 10 & 5 & 10 & 1.0 \\ \midrule[0.3pt]
Tolokers  & 1e-3 & 20 & 10 & 20 & 1.0  \\
Minesweeper  & 1e-3 & 20 & 10 & 20 & 0.1  \\
Amazon-ratings  & 5e-4 & 50 & 25 & 25 & 1.0  \\
Chameleon  & 1e-3 & 10 & 5 & 10  & 0.2  \\
\midrule[0.3pt]
\end{tabular}}}
\end{table}

\end{document}